\pdfoutput=1
\documentclass[10pt,twocolumn,letterpaper]{article}

\usepackage[pagenumbers]{cvpr} 
\usepackage{graphicx}
\usepackage[dvipsnames]{xcolor}
\usepackage{graphicx}

\definecolor{cvprblue}{rgb}{0.21,0.49,0.74}
\usepackage[pagebackref,breaklinks,colorlinks,citecolor=cvprblue]{hyperref}

\def\paperID{10305} 
\def\confName{CVPR}
\def\confYear{2024}

\title{DiffPortrait3D: Controllable Diffusion for Zero-Shot Portrait View Synthesis}

\author{Yuming Gu\textsuperscript{1,2}, You Xie\textsuperscript{2}, Hongyi Xu\textsuperscript{2},
Guoxian Song\textsuperscript{2}, Yichun Shi\textsuperscript{2}, \\
Di Chang\textsuperscript{1,2},{Jing Yang\textsuperscript{1}}, Linjie Luo\textsuperscript{2}\\
\textsuperscript{1}University of Southern California, \textsuperscript{2}ByteDance Inc. \\
\href{https://github.com/FreedomGu/DiffPortrait3D/}{https://freedomgu.github.io/DiffPortrait3D}\\
{\tt\small \{yuminggu,dichang,jyang010\}@usc.edu }\\
{\tt\small \{hongyixu,you.xie,guoxian.song,yichun.shi,linjie.luo\}@bytedance.com}
}

\begin{document}
\maketitle
\begin{abstract}
We present DiffPortrait3D, a conditional diffusion model that is capable of synthesizing 3D-consistent photo-realistic novel views from as few as a single in-the-wild portrait.  Specifically, given a single RGB input, we aim to synthesize plausible but consistent facial details rendered from novel camera views with retained both identity and facial expression. In lieu of time-consuming optimization and fine-tuning, our zero-shot method generalizes well to arbitrary face portraits with unposed camera views, extreme facial expressions, and diverse artistic depictions. At its core, we leverage the generative prior of 2D diffusion models pre-trained on large-scale image datasets as our rendering backbone, while the denoising is guided with disentangled attentive control of appearance and camera pose. To achieve this, we first inject the appearance context from the reference image into the self-attention layers of the frozen UNets. The rendering view is then manipulated with a novel conditional control module that interprets the camera pose by watching a condition image of a crossed subject from the same view. Furthermore, we insert a trainable cross-view attention module to enhance view consistency, which is further strengthened with a novel 3D-aware noise generation process during inference. We demonstrate state-of-the-art results both qualitatively and quantitatively on our challenging in-the-wild and multi-view benchmarks.
\end{abstract}

\section{Introduction}

\begin{figure}
	\centering
	\includegraphics[width=1.0\linewidth]{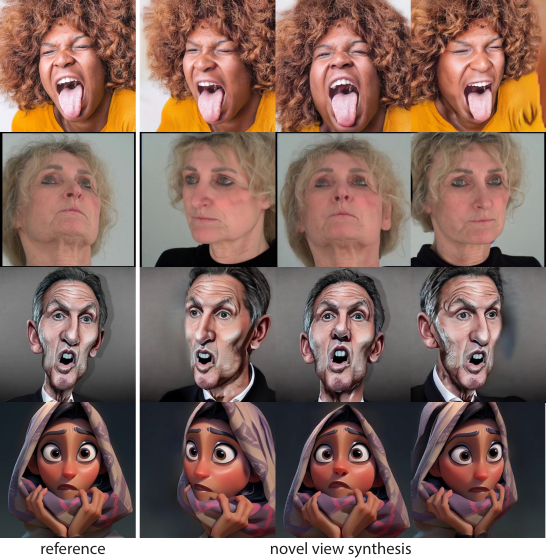}
	\centering
	\caption[Caption for LOF]{Given a single portrait as reference (left), DiffPortrait3D is adept at producing high-fidelity and 3d-consistent novel view synthesis (right). Notably, without any finetuning, DiffPortrait3D is universally effective across a diverse range of facial portraits, encompassing, but not limited to, faces with exaggerated expressions, wide camera views, and artistic depictions. }  
        \vspace{-4mm}
	\label{fig:teaser}
\end{figure}

Faithfully reconstructing the 3d appearance of human faces from a single  2D unconstrained portrait is a long-standing goal for computer vision, with a wide range of downstream applications in visual effects, digital avatars, 3D animation, and many others. 
In this work, we challenge ourselves to synthesize \emph{high-fidelity} \emph{consistent} novel views from as few as \empty{a single portrait}, with \emph{high resemblance} to the inputs in both individual appearance, expression and background content. Notably to the best of our knowledge, we are the first \emph{zero-shot} novel portrait synthesis work that supports versatile facial appearances and backgrounds, exaggerated expressions, wide views, and a plethora of artist styles. 

Long-range portrait view synthesis from sparse inputs requires a generative prior to hallucinating plausible scene features that are unobserved in the inputs. 
Recently, 3D aware generative adversarial network (GAN)~\cite{chan2021pi,gu2021stylenerf,or2021stylesdf,xue2022giraffe,Chan2022,epigraf,xu2021generative,deng2021gram, An_2023_CVPR} demonstrated striking quality and multi-view-consistent image synthesis, by integrating 3D neural representations~\cite{mildenhall2020nerf,xie2022neural} with style-based image generation~\cite{goodfellow2014generative,karras2019style,Karras2020stylegan2}.  Thereafter a line of work~\cite{yuan2023make,trevithick2023,bhattarai2023triplanenet,roich2022pivotal,lin20223d} has explored either optimization-based or encoder-based approaches to carefully invert the image into the latent or feature embedding of 3D GANs, and then synthesize novel views with 3D-aware generative priors. Nevertheless, almost all existing 3D-aware GANs are trained on limited image datasets. Hence when it comes to much more wild and nuanced portraits with large domain gap with the training distributions, GANs tend to struggle in faithfully depicting the 3D faces, resulting in loss of resemblance, corrupted geometry,  or blurry extrapolation (see Figure~ \ref{fig:compare_nvs_compare_style},~\ref{fig:compare_nvs_compare0}). 

With the recent advent of text-to-image diffusion models~\cite{ho2020denoising,song2020denoising,song2020score,rombach2022high}, we have witnessed unprecedented diversity and stability in image synthesis exhibited by large diffusion models pre-trained on billions of images, such as Imagen~\cite{saharia2022photorealistic} and Stable Diffusion (SD)~\cite{sd2022}. We therefore aim to capitalize on the generative power of production-ready diffusion models (SD in our work), for the task of portrait view synthesis. However, unlike previous 3D GAN-inversion works, simply inverting the reference image into a generative noise or a textual description does not naturally lift the image into a 3D scene, and it struggles to retain consistent appearances when deviating from the reference view. The introduction of ControlNet~\cite{zhang2023adding} enhances the controllability of Stable Diffusion by injecting localized spatial conditions. However, it remains unclear how to achieve appearance-disentangled view control such as in the paradigm of ControlNet.  Moreover,  without inherent 3D representation, the direct application of existing 2D image diffusion models to long-range animated view synthesis results in severe flickering artifacts. 

In this work, we propose \emph{DiffPortrait3D}, a novel zero-shot approach that lifts 2D diffusion model for synthesizing 3D consistent novel views from as few as a single portrait. Our key insight is to decompose the task into explicitly disentangled control of appearance and camera view. Specifically, we first utilize a trainable copy of the SD UNets to derive semantic appearance context from the reference image and then provide layer-by-layer contextual guidance to the self-attention modules of a locked SD network.  This allows us to preserve the capability of the large diffusion models while generating images with retained reference characteristics regardless of the rendering views.  On top of that, we further achieve view control by adding camera pose attention to the locked UNet decoder as done in ControlNet~\cite{zhang2023adding}. 
By design, the camera pose attention is intelligently extracted from an RGB portrait image of a proxy subject captured at the same view, to minimize appearance leakage from the condition image (e.g., shape and expression from landmarks). Additionally, to alleviate flickering artifacts when animating the views, we adopt a cross-view attention module as used in many video diffusion models~\cite{ho2022video,guo2023animatediff}. This ensures the unobservable region is completed in a consistent fashion.  
View consistency is further enhanced during inference with a novel 3D-aware noise generation process.

With the locked parameters of Stable Diffusion, we fine-tuned our control modules in stages with multi-view synthetic dataset by PanoHead~\cite{An_2023_CVPR} and real-image Nersemble dataset~\cite{kirschstein2023nersemble}.  Our method demonstrates native generalization capability to in-the-wild portraits without run-time fine-tuning. We extensively evaluate our framework on a few challenging benchmarks. DiffPortrait3D outperforms prior methods both quantitatively and qualitatively in terms of visual quality, resemblance, and view consistency.  
The contributions of our work can be summarized as:
\begin{itemize}
\item A novel zero-shot view synthesis method that extends 2D Stable Diffusion for generating 3d consistent novel views given as little as a single portrait.  

\item We demonstrate compelling fine-tuning-free novel view synthesis results given a single unconstrained portrait, regardless of its appearance, expression, pose, and style. 

\item Explicitly disentangled control of appearance and camera view, enabling effective camera control with preserved identity and expression. 

\item Long-range 3D view consistency with a cross-view attention module and 3D-aware noise generation.
\end{itemize}
Our code and model will be available for research purposes. 

\section{Related Works}
\begin{figure*}
   \centering
   \includegraphics[width=\linewidth]{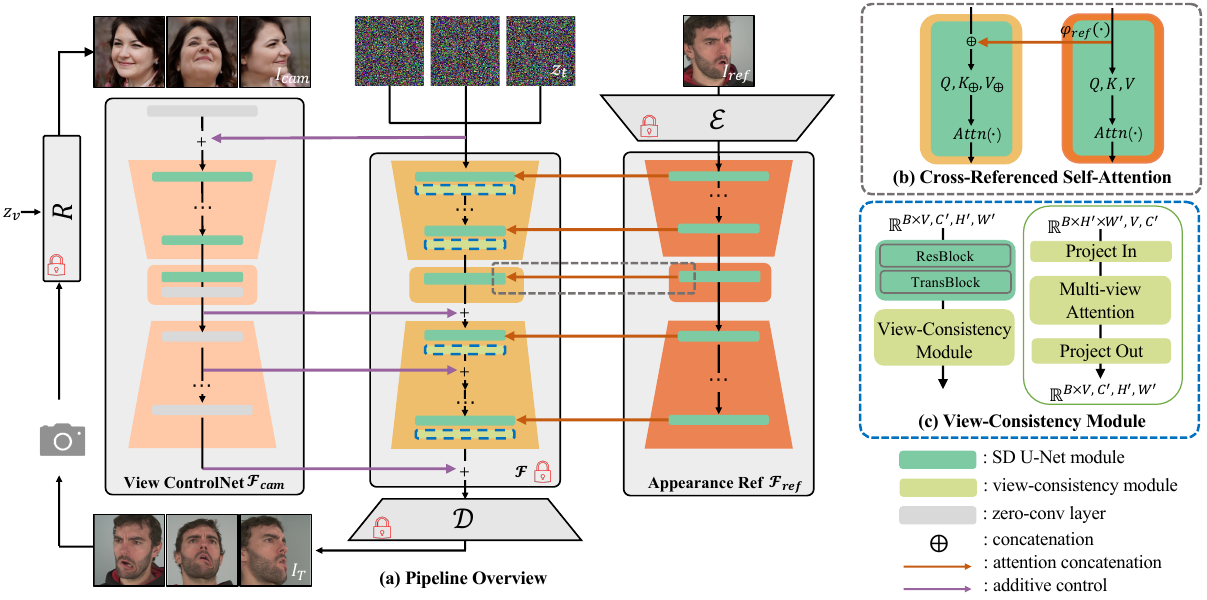}
   \caption{
   (a) Overview of our DiffPortrait3D framework. Given a single reference image $I_{ref},$ we aim to synthesize its novel views as $I_{T}$ at camera perspectives aligned with condition images $I_{cam}.$
   We leverage a pre-trained LDM $\mathcal{F}$ as our image synthesis backbone (middle), where its self-attention layers cross query the appearance context from $I_{ref}$ via our appearance reference module $\mathcal{F}_{ref}$ (right). Our view control module (left) $\mathcal{F}_{cam}$ derives additive view condition from $I_{cam}$ and exerts on $\mathcal{F}$. Additionally, we plug in view consistency modules (dotted rectangles, middle) to $\mathcal{F}$ to enhance multi-view coherence. During training, the images $I_{cam}$ are rendered using an off-the-shelf 3D GAN renderer $R$, where its camera perspectives are aligned with $I_{T}$. (b) The intermediate spatial features $\varphi(\cdot)$ sourced from $I_{ref}$ are concatenated into the corresponding self-attention blocks in $\mathcal{F}$. 
   (c) An attention mechanism is employed across the multi-view dimensions by our view-consistency module. 
   }
   \vspace{-5mm}
   \label{fig:architecture}
\end{figure*}
Our study focuses on the application of 2D diffusion models for zero-shot portrait novel view synthesis (NVS). Within this context, we undertake an extensive survey of progress in techniques related to novel view synthesis, categorized into regression-based and generative approaches. 
\vspace{-3mm}
\paragraph{Regression based NVS.} 
Facial NVS is attainable through the use of explicit parametric geometry priors, as demonstrated by 3D Morphable Models (3DMM)~\cite{paysan20093d, tsutsui2022novel, wu2019mvf, dib2021high, zhuang2022mofanerf}.
However, the limited parametric space of 3DMM poses challenges in faithfully depicting diverse facial expressions.
Recent strides in Neural Radiance Fields (NeRF)~\cite{mildenhall2020nerf,gu2021stylenerf,hong2022headnerf, zhang2022fdnerf} have yielded high-fidelity results in novel view synthesis.
Notably in the realm of portrait NVS, FDNeRF~\cite{zhang2022fdnerf} constructed a NeRF model that integrates aligned features from inputs to generate novel view portraits. Nevertheless, achieving photo-realistic 3D-aware novel views with such models typically necessitates the availability of dense calibrated images.
\vspace{-3mm}
\paragraph{Generative NVS with GAN}
GANs~\cite{goodfellow2014generative} employ adversarial learning to synthesize images that faithfully capture the distribution of the training dataset. Previous studies have demonstrated the effectiveness of 2D GANs in portrait manipulation, employing techniques such as latent space exploration~\cite{chen2016infogan} and exemplar image utilization~\cite{kafri2021stylefusion, xiang2020one}. Nevertheless, the absence of inherent 3D representations in these 2D GANs presents a challenge in maintaining 3D consistency for the task of NVS.

Recent advancements on 3D aware GANs~\cite{chan2021pi,gu2021stylenerf,or2021stylesdf,xue2022giraffe,Chan2022,epigraf,xu2021generative,deng2021gram, An_2023_CVPR}, built upon foundations of 2D GANs, have demonstrated 
striking quality and multi-view-consistent image synthesis.
These methodologies typically leverage StyleGAN2~\cite{karras2020analyzing} as a fundamental component, incorporating it with differential rendering and diverse 3D representations, such as signed distance functions as in StyleSDF~\cite{or2021stylesdf} and tri-plane representations used by EG3D~\cite{Chan2022}. 
 Thereafter a line of work~\cite{yuan2023make,trevithick2023,bhattarai2023triplanenet,roich2022pivotal,lin20223d} has explored either optimization-based or encoder-based approaches to carefully invert the image into the latent or feature embedding of 3D GANs, and then synthesize novel views with 3D-aware generative priors. 
It is noteworthy, however, that these methods heavily depend on a pre-trained 3D GAN generator and exhibit limitations in their capacity to generate unposed portraits with in-the-wild expressions, styles, and camera views. 
\vspace{-3mm}
\paragraph{Diffusion Model based NVS} 
In lieu of directly confronting the intricacies of learning a 3D diffusion model, recent research endeavors have embraced an alternative strategy, harnessing powerful 2D diffusion models to improve the processes of 3D modeling and novel view synthesis. DreamFusion~\cite{poole2022dreamfusion} pioneered this strategy by distilling a 2D text-to-image generation model for fine-tuning a NeRF model. GENVS~\cite{chan2023genvs} introduced a diffusion-based model explicitly tailored for 3D-aware generative novel view synthesis from a single input image. 
Their methodology involves modeling samples from the potential rendering distribution, effectively mitigating ambiguity and generating plausible novel views through the utilization of diffusion processes. 
Recent noteworthy study, Zero-1-to-3~\cite{liu2023zero1to3, liu2023one2345} 
utilizes a stable diffusion model to capture geometric priors derived from an extensive synthetic dataset, yielding high-quality predictions. 
Moreover, Consistent123~\cite{lin2023consistent123}, a case-aware approach, utilizes Zero-1-to-3 as 3D prior for the initial structural representation before generating high texture fidelity.
However, it is crucial to note that these approaches primarily concentrate on general objects, resulting in a diminished quality when applied to portrait synthesis.
\section{Methods}

\begin{figure*}
	\centering
        \vspace{-4mm}
	\includegraphics[width=0.95\linewidth]{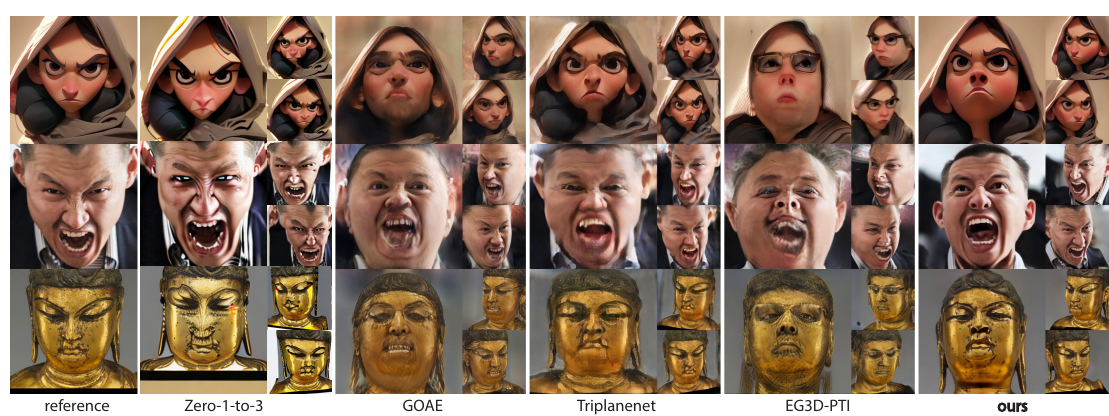}
	\centering
	\caption[Caption for LOF]{\textbf{Qualitative comparison of novel view synthesis on in-the-wild images.} 
 Compared to the baselines, our method shows superior generalization capability to novel view synthesis of wild portraits with unseen appearances, expressions and styles, even without any reliance on fine-tuning. 
 }  
 \vspace{-3mm}
	\label{fig:compare_nvs_compare_style}
\end{figure*}
Given as few as a single RGB portrait image, denoted as $I_{ref}$, captured from any camera perspective, we aim to synthesize a new image $I_{T}$ at a novel query view as indicated by a condition image $I_{cam}$. The synthesized image $I_{T}$ should retain the expression and appearance of the foreground individual as well as the background context as in $I_{ref},$ while follows the rendering view of $I_{cam}.$ Note that $I_{cam}$ and $I_{ref}$ could be of a completely different identity. 

Our proposed approach, DiffPortrait3D, leverages a latent diffusion model (LDM) as the backbone of our rendering framework, as depicted in Figure~\ref{fig:architecture} (a) (Section~\ref{sec:pre}). 
We then introduce an auxiliary appearance control branch (Section~\ref{sec:app}) to exert layer-by-layer guidance with local structures and textures  from reference images $I_{ref}.$ To enable effective camera control with $I_{cam},$ our view control module, designed in a fashion of ControlNet~\cite{zhang2023adding}, implicitly derives camera pose from $I_{cam}$ and inject to the diffusion process as an additive condition (Section~\ref{sec:view}). Lastly we discuss about enhancing view consistency with our integrated multi-view attentions, and noise generation with 3D awareness at inference (Section~\ref{sec:temporal}).  

\subsection{Preliminaries}
\label{sec:pre}
\paragraph{Latent Diffusion Models.} Diffusion models~\cite{ho2020denoising,song2020denoising,song2020score} are generative models designed to synthesize desired data samples from Gaussian noise via removing noises iteratively.
Latent diffusion models~\cite{rombach2022high} are a class of diffusion models that operates in the encoded latent space of an autoencoder $\mathcal{D}(\mathcal{E}(\cdot)),$ where $\mathcal{E}$ and $\mathcal{D}$ denotes the encoder and decoder respectively. Specifically, given an image $I$ and the text condition $c_{text}$, the encoded image latent $z_0$ = $\mathcal{E}(I)$ is diffused
$T$ time steps into a Gaussian-distributed $z_T \sim \mathcal{N}(0, 1)$. 
The model is then trained to learn the reverse denoising process with the objective,
\begin{equation}
    L_{ldm} =\mathbb{E}_{z_0,c_{text},t,\epsilon \sim \mathcal{N}(0,1)} \bigg[ \Big\lVert \epsilon-\epsilon_\theta \big(z_t, c_{text},t\big) \Big\lVert_2^2 \bigg],
\end{equation} 
The $\epsilon_\theta$ is formulated as a trainable U-Net architecture with layers of intervened convolutions (ResBlock) and self-/cross-attentions (TransBlock). In this paper, we build our network as a plug-and-play module to the recent state-of-the-art text-to-image latent diffusion model, Stable Diffusion~\cite{sd2022}.   
\vspace{-3mm}
\paragraph{ControlNet.} As introduced by~\cite{zhang2023adding}, ControlNet effectively enhances latent diffusion models with
spatially localized, task-specific image conditions.
As its core, it replicates the original Stable Diffusion as a trainable side path, and adds additional ``zero convolution'' layers. The extra conditions outputted from the ``zero convolution'' layers are then added to the skipped connections of the SD-UNets.  Let $c_p$ be the extra condition,
the noise prediction of U-Net with ControlNet then becomes
$\epsilon_\theta \big(z_t, c_{text},c_{p}, t\big)$.   

\vspace{-1.5mm}
\begin{figure*}
	\centering
	\includegraphics[width=0.97\linewidth]{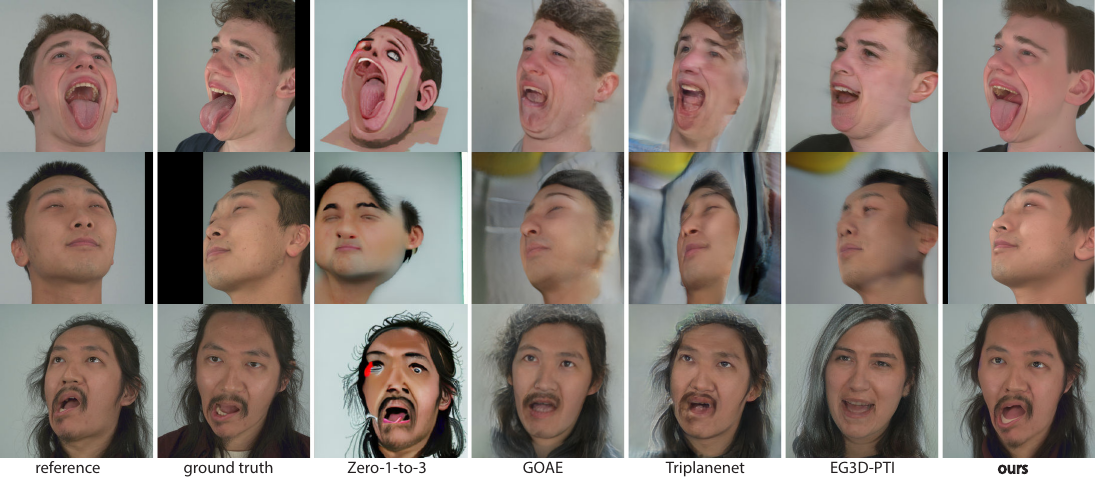}
	\centering
	\caption[]{\textbf{Qualitative comparison of novel view synthesis on NeRSemble~\cite{kirschstein2023nersemble}.}
 Our method achieves effective view control for novel synthesis with the best perceptual quality and retained identity and expression, even for portraits with exaggerated expressions and under substantial change of camera view for synthesis. }  
        \vspace{-5mm}
	\label{fig:compare_nvs_compare0}
\end{figure*}

\subsection{Appearance Reference Module}
\label{sec:app}
In order to synthesize a novel view of $I_{ref}$ with LDM, one could try to condition the denoising with an ``inverted'' text condition $c_{text}$~\cite{li2023blip}.  
However, providing a precise textual description of $I_{ref}$ for LDM to comprehensively recover all its components is often a challenging undertaking. 
Alternatively, one could also condition $\epsilon_{\theta} $ on $I_{ref}$ directly as a ControlNet. Such a design, however, tend to generate images predominantly influenced by the camera pose in $I_{ref}.$
Inspired by~\cite{cao2023masactrl,weng2023consistent123}, we opt for integrating appearance attributes of the reference image $I_{ref}$ into the UNet backbone as cross-referenced self-attentions. 
Note that to eliminate the harmful influence of inaccurate text description, we set $c_{text}$ empty and use the reference image $I_{ref}$ as the only source of appearance.



To illustrate our appearance reference module, 
let us denote the pretrained LDM as $\mathcal{F}$,
where its self-attention is calculated as 
\begin{gather}
    Attn(\cdot) = softmax(\frac{QK^T}{\sqrt{d}})\cdot V\\
   Q=W_Q\cdot \mathcal \varphi(z_t), K=W_K \cdot \mathcal \varphi(z_t), V=W_V\cdot \mathcal \varphi(z_t),
\end{gather}
where $Q, K, V$ are the query, key, and value features projected from the spatial features $\varphi(z_t)$ with corresponding projection matrices respectively.

To guide the denoising process with $I_{ref},$ we adapt the self-attention mechanism within $\mathcal{F}$ such that it is able to cross query the correlated local contents and textures from $\mathcal{E}(I_{ref})$, in addition to its own spatial features. 
Specifically we replicate $\mathcal{F}$ into a trainable counterpart $\mathcal{F}_{ref}$ with $\varphi_{ref}(\cdot)$ serving as intermediate representations within the UNet architecture.
As depicted in Figure \ref{fig:architecture} (b), we then modify the vanilla self-attention in $\mathcal{F}$ in a way that the spatial context $\varphi_{ref}(\mathcal{E}(I_{ref}))$ in the appearance branch $\mathcal{F}_{ref}$ is cross-queried layer by layer as,  
\begin{equation}
    \begin{aligned}
        K_{\oplus}&=W_K \cdot (\varphi(z_t) \oplus \varphi_{ref}(\mathcal{E}(I_{ref}))), \\
        V_{\oplus}&=W_V\cdot (\varphi(z_t) \oplus \varphi_{ref}(\mathcal{E}(I_{ref}))),
    \end{aligned}
\end{equation}
where $\oplus$ denotes concatenation. 
Note that we do not apply noise to $I_{ref},$ 
ensuring meticulous transfer of referenced structure and appearance attributes into the novel portrait synthesis. 
We lock the parameters of SD-UNet $\mathcal{F}$, and train our appearance reference module $\mathcal{F}_{ref}$ with paired multi-view images.  

Notable, when more reference images are available, e.g., in some multi-view capture settings, our appearance reference module can be easily extended by concatenating multiple appearance contexts as
\begin{equation}
    \begin{aligned}
        \varphi(z_t) \oplus \varphi_{ref}(\mathcal{E}(I^1_{ref})) \oplus ... \oplus \varphi_{ref}(\mathcal{E}(I^n_{ref})).
    \end{aligned}
\end{equation}
Our trained module is capable of seamlessly integrating the multi-view appearance clues into 3D-consistent appearance context (Figure~\ref{fig:multi_src}).



\subsection{View Control Module}
\label{sec:view}
In this stage, we aim to attain control over the synthesis viewpoint without influencing either the derived appearance attributes by $\mathcal{F}_{ref}$ or the synthesis capability of a pre-trained LDM $\mathcal{F}$.  This naturally leads to the paradigm of ControlNet~\cite{zhang2023adding} where the additional view control is connected via ``zero convolution'' layers of a trainable LDM copy, with both $\mathcal{F}_{ref}$ and $\mathcal{F}$ locked. Here we denote our view control module as $\mathcal{F}_{cam},$  to be trained with multi-view images. One straightforward design of $\mathcal{F}_{cam}$ would be to employ the spatial feature maps extracted from the ground-truth target images as image conditions, such as landmarks, segmentation, or edges. We note that such ``ground-truth'' condition images are not available during inference and therefore the view is typically manipulated with images of a different identity. 
However, we argue that such condition images contain entangled semantic appearance information, such as shape and expression, which is likely to be passed along with the camera pose to $\mathcal{F}$. Herein, appearance leakage from the view condition image will be reflected on the novel view synthesis during inference. This artifact is more pronounced when $I_{ref}$ and $I_{cam}$ exhibit distinct appearance features. 

\begin{figure}
	\centering
	\includegraphics[width=0.97\linewidth]{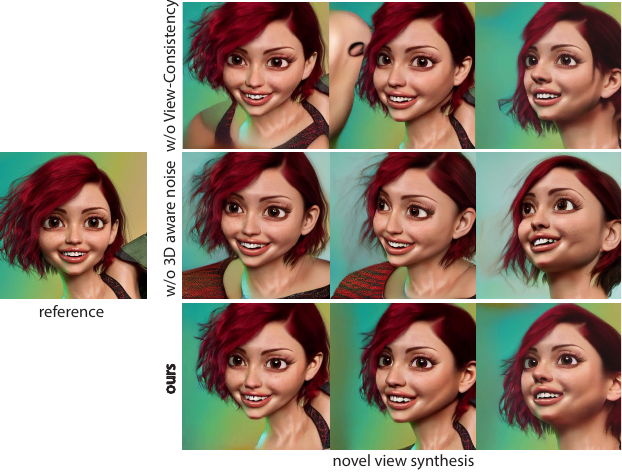}
	\centering
        \vspace{-3mm}
	\caption[Caption for LOF]{\textbf{Ablation on view consistency.} Excessive background variation and slight shading change across multiple novel views are observable without our view-consistency module. Our 3D-aware noise, compared to random Gaussian noise, helps maintain structural coherence during view animation. }  
        
	\label{fig:alba_MM}
\end{figure}

Instead, we utilize a portrait image from a distinct random identity as the view condition, and generate novel-view images that mirror the head pose as in the condition portrait $I_{cam}$. Our design unifies the view manipulation setting in training and inference, 
and facilitates the natural disentanglement of view and appearance control. 
However, training ControlNet for cross-identity view control requires paired images at a  identical view, and obtaining such data pairs is typically unfeasible in real-world capture settings. To address this hurdle, we leverage off-the-shelf 3D GAN renders $\mathcal{R} (v, z_v)$, as exemplified in prior works \cite{Chan2022, An_2023_CVPR}, to generate synthetic pose images $I_{cam}$. Here the $v$ denotes the camera parameters calibrated from the target image and $z_v$ is a random Gaussian noise input to the 3D GAN. Since $I_{cam}$ and $I_{ref}$ possess substantial difference in expression and appearance, our view control module is therefore instructed to derive camera pose from $I_{cam}$ only. Moreover, by design, the camera pose is directly interpreted by our view control module, allowing us to mimic the rendering view simply with an RGB image. This largely eases the cumbersomeness in feature processing of $I_{cam},$ e.g., landmarks detection or semantic parsing, which could be unreliable with heavy occlusion or under wide views. 
\subsection{View Consistency Module}
\label{sec:temporal}
To this end, we have facilitated the generation of a novel-view portrait via the seamless combination of an appearance reference module,  a view ControlNet and a pre-trained LDM. Nevertheless, achieving consistency in features across various views poses a significant challenge as many explanations exist for the unobservable region. Inspired by AnimateDiff \cite{guo2023animatediff}, we introduce a view consistency module that incorporates cross-view attention within a batch of views. 
Such a module employ an attention mechanism along the dimension of views to establish feature correlation among the multiple novel view synthesis. Similar to AnimateDiff, we integrate these view consistency modules into the up- and down-sampling blocks of the LDM $\mathcal{F}$, as depicted in Figure \ref{fig:architecture} (c). 
However, we note that such frame-wise modules were originally proposed for temporal coherence and as motion prior, trained with sequential video frames. In contrast, the animated view motion is purely defined by the sequence of $I_{cam}.$ Therefore, we trained our view-consistency modules with batches of randomly shuffled views, permitting the modules to focus on cross-view attentions in lieu of motion distribution.

 \begin{table*}[t]
 	\renewcommand{\tabcolsep}{2pt}
 	\small
 	\begin{subtable}[!t]{0.6\textwidth}
		\centering
\begin{tabular}{ccccc}
    \toprule
     &Ours
     &Eg3D-PTI
     &GOAE
     &Triplanenet
     \\
    \midrule
    POSE $\downarrow$ & -/0.0068/-   & -/\textbf{0.0021}/- & -/0.0022/- & -/0.0134/- \\ 
    \midrule        
    LPIPS $\downarrow$ & \textbf{0.02}/\textbf{0.23}/\textbf{0.02} & 0.21/0.26/0.36 & 0.12/0.28/0.21 & 0.10/0.39/0.17\\ 
    \midrule
    SSIM $\uparrow$ & \textbf{0.92}/\textbf{0.62}/\textbf{0.93} & 0.66/0.59/0.46 &  0.72/0.57/0.54 & 0.76/0.50/0.63   \\
    \midrule
    DIST $\downarrow$ & \textbf{0.06}/\textbf{0.21}/\textbf{0.04} & 0.21/0.24/0.27 & 0.15/0.25/0.20 &  0.15/0.29/0.19 \\    
    \midrule    
    ID $\uparrow$ & \textbf{0.95}/\textbf{0.70}/\textbf{0.92} & 0.15/0.28/0.12  & 0.55/0.39/0.54 & 0.70/0.45/0.70 \\ 
    \midrule
    FID $\downarrow$ & \textbf{7.65}/\textbf{27.4}/\textbf{11.7} & 33.13/56.2/95.0 & 54.10/84.8/92.0 & 63.54/112.1/88.0 \\
   \bottomrule
\end{tabular}
\caption{}
\label{tab: nvs}
\end{subtable}
\begin{subtable}[!t]{0.4\textwidth}
		\centering
\begin{tabular}{cccc}
    \toprule

    &w/o  $\mathcal{F}_{ref}$  & $I_{ref}$ & $I_{ref}$\\
        & finetuning & unaligned &  aligned \\
    \midrule
    LPIPS $\downarrow$ & 0.55& 0.28& \textbf{0.27}\\ 
    \midrule
    SSIM $\uparrow$ & 0.47& \textbf{0.68} & \textbf{0.68}\\
    \midrule
    DIST $\downarrow$ & 0.43&0.19 &\textbf{0.18}\\    
    \midrule    
    ID $\uparrow$ & 0.21&\textbf{0.70}&\textbf{0.70}\\
    \midrule
    FID $\downarrow$ & 99.09 &25.77& \textbf{25.37}\\
   \bottomrule
\end{tabular}
\caption{}
\label{tab:app}
\end{subtable}
    \caption{(a) Quantitative comparison of our method and GAN-based baselines, showing numerical results of reconstruction/novel view synthesis of NeRSemble~\cite{kirschstein2023nersemble}, and reconstruction of in-the-wild test images( from left to right). For a fair comparison to our baselines, the evaluation is performed at the resolution of $256\times 256$.  
     (b) Ablation study of our method without finetuning appearance reference module, with unaligned reference images, and with aligned reference images, evaluated on NeRSemble at the resolution of $512\times 512$.}\label{tab:second}
     \vspace{-3mm}
 \end{table*}
 
As illustrated in Figure~\ref{fig:architecture} (c), we train our view-consistency modules in groups of multi-view condition images $\{I_{cam}\}$ with a shape of $(B, V, C, H, W)$, where $B$ and $V$ are the batch size and the number of views, while $C, H, W$ denote the number of image channels, image height and width respectively. We note that the appearance within each batch of generation is referenced from the same image $I_{ref}$. 
Inside $\mathcal{F}$, we reshape the input to ResBlocks and TransBlocks as $(B \times V, C', H', W')$, where $C', H', W'$ represent the latent feature channel, height and width respectively. 
Following the operations of self- and cross-attention, we then transform the layer input into a shape of $(B \times H' \times W', V, C')$, performing view-wise attention within the view consistency modules. 
\begin{figure}[!t]
	\centering
	\includegraphics[width=0.97\linewidth]{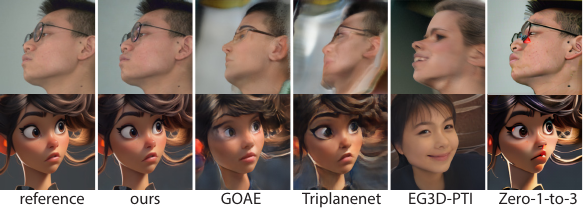}
	\centering
	\caption[]{ \textbf{Reconstruction}.
 DiffPortrait3D demonstrates meticulous reconstruction of referenced appearance, even with side views and 3D cartoon styles, substantially outperforming the baseline methods. }  
        \vspace{-3mm}
	\label{fig:rec_compare0}
\end{figure}
\paragraph{3D-aware inference.} It has been empirically observed that the image layout is formed in the early denoising steps. Therefore instead of denoising from multiple random Gaussian noises,  structural and textural consistency is likely to be enhanced when synthesizing multiple novel views by initiating the denoising process from ``3D-consistent'' noise samples. We propose an efficient two-stage process to generate noise samples with 3D awareness. On our multi-view image dataset, we first trained a 3D-convolution based NVS model with inclusion of 3D feature field and neural feature rendering (please refer to the supplementary paper for details). We employ this NVS model to provide a proxy synthesis $\tilde{I}$ at the target novel view, which is typically blurry but 3D consistent. We then diffuse the latent feature $\mathcal{E}(\tilde{I})$ with 1000 time steps into a Gaussian noise as the input to the LDM. In essence, the two-step generated noise still contains some image layout semantics 
in a very coarse grain and in practice, enhanced view consistency is observed in our task as demonstrated in Figure~\ref{fig:alba_MM}.

\section{Experiments}



\paragraph{Dataset and Training.} 
Our model was trained in three stages on our multi-view image dataset as an image reconstruction task. That being said, both the appearance reference image $I_{ref}$ and the target image $I_{T}$ are sourced from the same identity but with different views, whereas $I_{cam}$ is synthesized with EG3D~\cite{Chan2022} using a random latent Gaussian noise and the calibrated camera parameters of $I_{T}$. We lock the parameters of the SD-UNet $\mathcal{F}$ during the whole training stage. In the first stage, we train all the parameters of our appearance reference module $\mathcal{F}_{ref}$ without any camera guidance. Next we freeze the weights of $\mathcal{F}_{ref},$ and train our view control module $\mathcal{F}_{cam}$ with paired $I_{cam}.$ Lastly the view consistency module, performing cross-view attentions among 8 views at once, is trained with the rest modules frozen. 
All training was conducted on 6 Nvidia A100 GPUs at a learning rate of $10^{-5}$, with 16 images processed in each step. During inference, we empirically set 100 steps for DDIM denoising~\cite{song2020denoising} and unconditional guidance scale~\cite{ho2022classifier} as 3 for a good balance of quality and speed.

We trained our modules on a hybrid dataset comprised of photo-realistic multi-view images NeRSemble~\cite{kirschstein2023nersemble} and synthetic ones by PanoHead~\cite{An_2023_CVPR}. 
NeRSemble dataset consists of high-resolution videos of 220 subjects performing a wide range of dynamic expressions, captured from 16 calibrated synchronized cameras. We sampled 2000 pairs of multi-view frames from NeRSemble for training, where 1 randomly-selected view is used for appearance reference and 8 other views as targets. Given the scarcity of available camera views and the background variation, we augmented our training dataset with another 2000 pairs of multi-view images synthesized via PanoHead~\cite{An_2023_CVPR}. 
For evaluation, we used another unseen 500 multi-view pairs from NeRSemble, and 360 single-view internet-collected in-the-wild portraits, containing a wide variation in appearance, expression, camera perspective, and style. 
We note that for training, all the images are cropped and aligned as in EG3D~\cite{Chan2022} whereas we do not perform image alignment during inference (unless explicitly stated for comparison to GAN-based methods). For testing on both datasets, the novel camera views are all manipulated with EG3D renderings. 



\subsection{Qualitative Evaluations}
Given a single reference portrait, our method demonstrates high-fidelity and 3D-consistent novel view synthesis at a resolution of $512\times 512,$ as illustrated in Figure~\ref{fig:teaser}. While only being trained on aligned real portrait images, our method shows superior generalization capability to novel identities, styles, expressions and views. This is largely credited to the preservation of the generative prior of pre-trained LDM by our design. As evidenced in Figure~\ref{fig:compare_nvs_compare0}, our view control module is also able to effectively control the synthesis view.  Compared to the ground truth (second column, Figure~\ref{fig:compare_nvs_compare0}), our novel portraits are highly plausible but with some noticeable  identity differences. This is due to the limited visual  appearance clue in the single reference image, and the problem can be largely alleviated with additional references (please refer to Figure~\ref{fig:multi_src} and the supplementary paper for visual results). 

We extensively compare to a few state-of-the-art novel portrait synthesis works on both image reconstruction (Figure~\ref{fig:rec_compare0}) and novel view synthesis (Figure~\ref{fig:compare_nvs_compare_style},~\ref{fig:compare_nvs_compare0}): GOAE~\cite{yuan2023make},TriPlaneNet~\cite{bhattarai2023triplanenet}, Pivot Tuning (EG3D-PTI) \cite{Chan2022} and Zero-1-to-3~\cite{liu2023zero1to3}. GOAE~\cite{yuan2023make} and TriPlaneNet~\cite{bhattarai2023triplanenet} designed an effective image encoder for EG3D~\cite{Chan2022}, whereas EG3D-PTI runs latent code optimization and finetunes the weights of EG3D per image. We did not compare to Live3D Portrait~\cite{trevithick2023realtime} given unavailable implementation and model. Zero-1-to-3~\cite{liu2023zero1to3} leverages Stable Diffusion but was trained on 3D object dataset Objaverse~\cite{deitke2023objaverse}. While not required by our method, we cropped and aligned the test images as in EG3D. Nevertheless, our method outperforms substantially over the prior work in terms of both perceptual quality, and preservation of identity and expression. Notably all 3D GAN-based baselines fail to reconstruct side views (Figure~\ref{fig:rec_compare0}),  exaggerated expressions (Figure~\ref{fig:compare_nvs_compare0}), or out-of-domain styles (Figure~\ref{fig:compare_nvs_compare_style}), whereas Zero-1-to-3 synthesizes novel portraits with very limited perceptual quality.


\subsection{Quantitative Evaluations}
We evaluate methods for single-view novel portrait synthesis on 4 main aspects. We use LPIPS$\downarrow$~\cite{zhang2018unreasonable},
DISTS$\downarrow$~\cite{ding2020iqa}, SSIM$\uparrow$~\cite{wang2004image} for evaluation of 2D image reconstruction, ID$\uparrow$~\cite{deng2019arcface} for identity consistency, FID$\downarrow$~\cite{heusel2017gans} for perceptual quality, and POSE $\downarrow$ for camera view control accuracy. Notably, to evaluate reconstruction fairly, we estimate camera parameters from the ground-truth target image and uses the EG3D renderings as condition $I_{cam}.$ The error in camera estimation could result in some image misalignment and therefore we mainly rely on perceptual metrics LPIPS and DISTS for reconstruction evaluation. The identity similarity is calculated 
between the synthesized and reference image by calculating the cosine similarity of the face embeddings with a pretrained face recognition module~\cite{deng2019arcface}. 

Table~\ref{tab: nvs} shows the numerical comparison on 
reconstruction of NeRSemble and in-the-wild test images, and novel view synthesis of NeRSemble respectively. On all image metrics, our method shows  our method is superior than all prior work by a large margin, demonstrating the most compelling image quality. Our pose reconstruction is slightly worse than the baseline. However, we argue that this is largely due to the camera misalignment between the ground truth and the condition EG3D rendering.

\subsection{Ablations}
\begin{figure}
	\centering
	\includegraphics[width=0.97\linewidth]{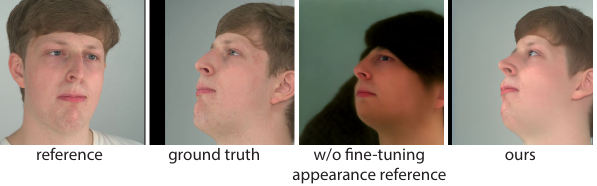}
	\centering
	\caption[Caption for LOF]{Fine-tuning appearance reference module helps better retain the spatial features from the reference image. }  
        
	\label{fig:alba_appear}
\end{figure}

\begin{figure}
	\centering
	\includegraphics[width=0.97\linewidth]{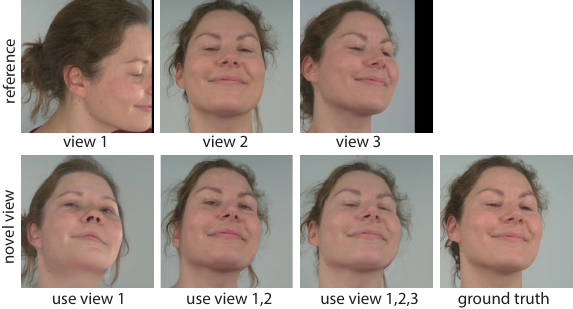}
	\centering
	\caption[Caption for LOF]{Our method seamlessly supports multiple reference images as input, and the novel view synthesis quality is progressively enhanced with more references. }  
        \vspace{-5mm}
	\label{fig:multi_src}
\end{figure}

We ablate the efficacy of the individual component with extensive ablation experiments for noval view synthesis on NeRSemble test set. 
As illustrated in Figure~\ref{fig:alba_MM}, we demonstrate the necessity of our view consistency module and 3D-aware noise in maintaining appearance coherence cross multiple views. 
Without them, substantial variations are observed, especially on the unobserved region of the reference image,  when altering the camera views. The weights of our appearance reference module is initiated from a copy of SD-UNet which should be already able to derive local appearance context from the reference image. However, as evidenced by Table~\ref{tab:app} and Figure~\ref{fig:alba_appear}, significant improvements are achieved by our finetuning on multi-view images. We reason that the necessity of finetuning is due to the removal of cross attention from text. 
Lastly unlike many GAN-based methods that requires the reference image to be aligned, our model supports free-form portraits as inputs without  quality degeneration even though the model was trained on camera-aligned multi-view images.  This is numerically shown in Table~\ref{tab:app} where aligning the reference image (as in EG3D) only leads to neglectable differences.

\section{Discussion}
\paragraph{Conclusion.} We presented \emph{DiffPortrait3D}, a novel conditional diffusion model that is capable of generating consistent novel portraits from sparse input views. By design, our framework seamlessly cross-references the key characteristics from the input images and effectively adds camera pose control into the latent diffusion process, modulated with enhanced consistency across views. Trained only with a few thousand of synthetic and real multi-view images, our model successfully showcases compelling novel portrait synthesis results, regardless of appearances, expressions, camera perspectives, and styles. This is largely credited to our explicitly disentangled control of appearance and view within both model design and training, without harming the generalization capability of large pretrained diffusion models.  We believe that our framework
opens up possibilities for accessible 3D reconstruction and visualization from a single picture. 

\clearpage
\setcounter{page}{1}
\maketitlesupplementary
\appendix

In this supplementary paper, we provide additional implementation details in Section~\ref{sec:3Dvolume}, showcase more visual results and numerical comparisons in Section~\ref{sec:moreres}, and discuss limitations \& ethics consideration in Section~\ref{sec:limit} and Section ~\ref{sec:ethic consider} . 

\section{Implementation Detail}
\label{sec:3Dvolume}
\begin{figure}
	\centering
	\includegraphics[width=\linewidth]{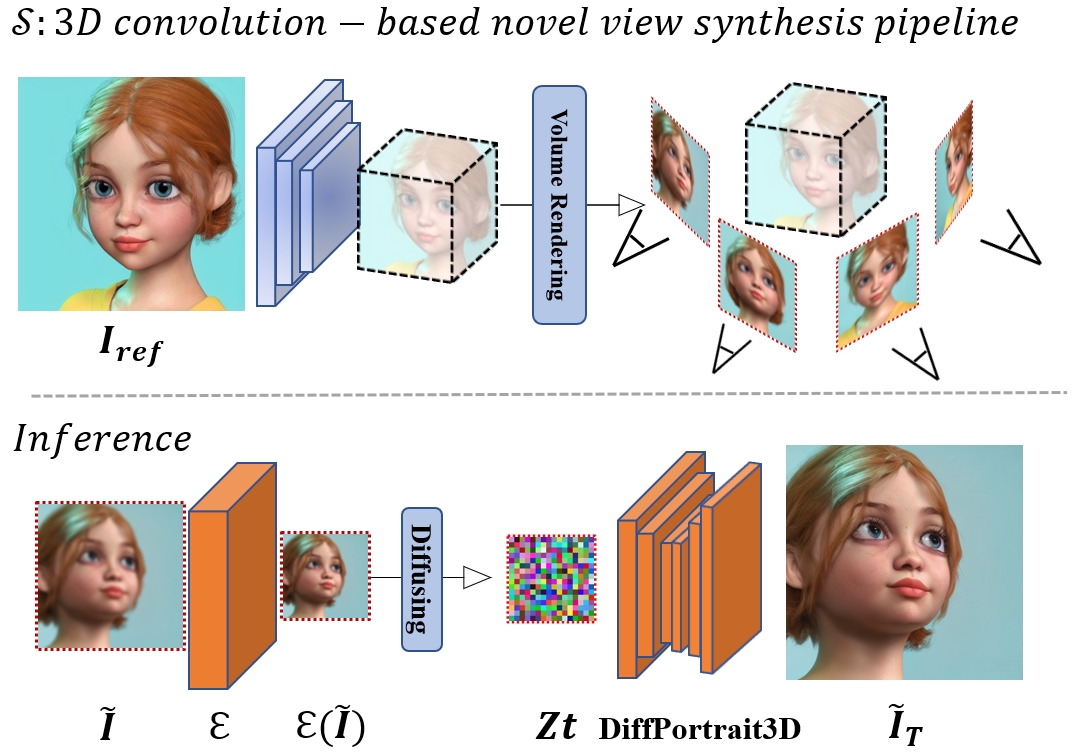}
	\centering
	\caption[Caption for 3D volume rendering]{shows how our 3D convolution-based novel view synthesis pipeline \textbf{$\mathcal{S}$} works. In practice, a 3D-convolution-based network first maps the reference image into a 3D feature volume. Then, given a conditioned camera view, we follow the volume rendering to integrate the 3D features into a 2D feature map which is further decoded to the final RGB image $\tilde{I}$ with a 2D convolution network. During the inference phase, we enhance 3D awareness by commencing with noise generated via a 1000-step forward diffusion process applied to \(\mathcal{E}(\tilde{I})\), which serves as the initial noise for our DiffPortrait3D pipeline.
 }  
	\label{fig:3dnoise}
\end{figure}
\subsection{3D-Aware Noise}
\begin{figure}
	\centering
        
	\includegraphics[width=\linewidth]{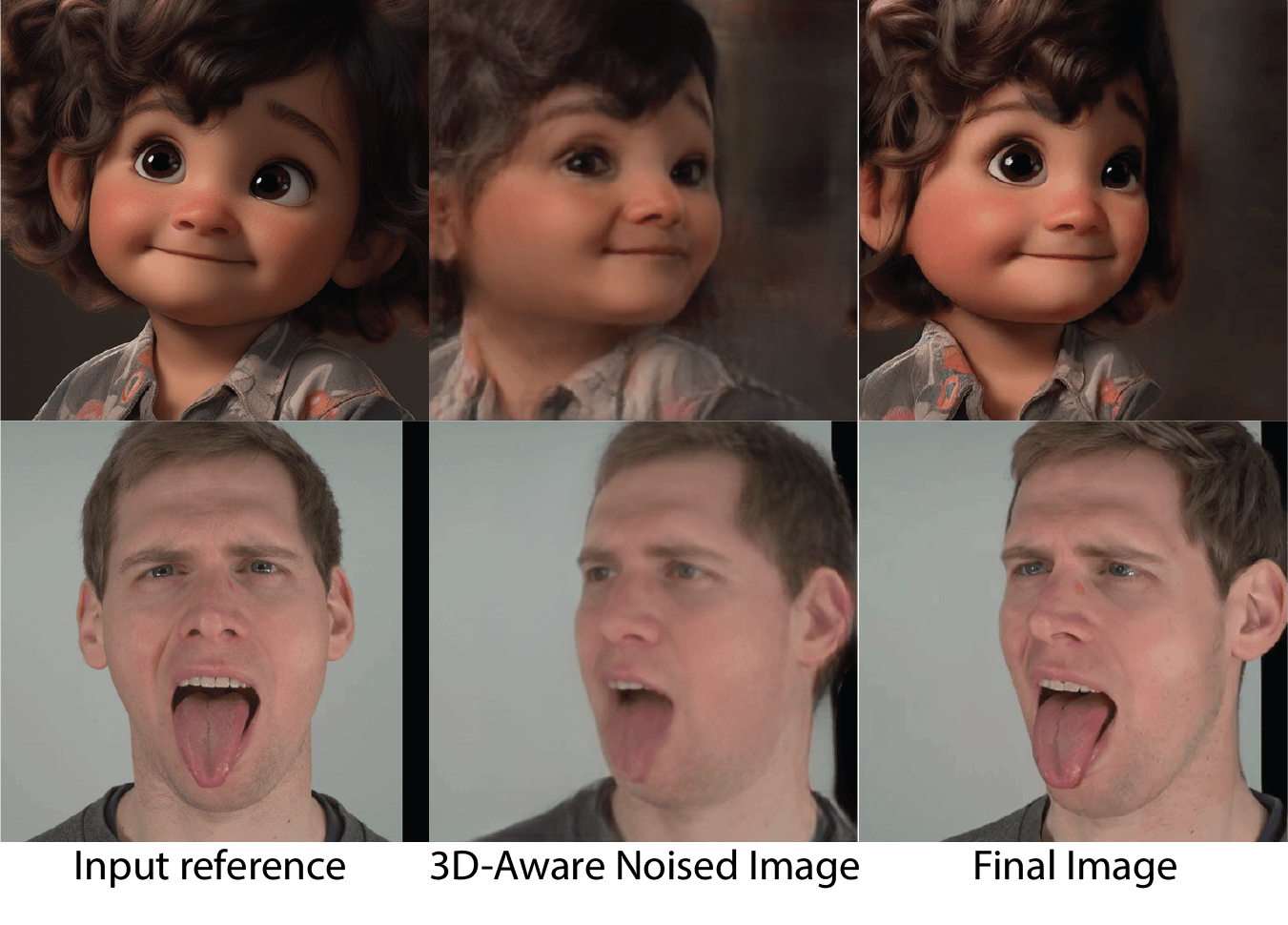}
	\centering
	\caption[Caption for 3D volume rendering]{Our 3D-Aware Noise effectively helps strengthen the novel view synthesis result.
 }  
	\label{fig:3dnoise_vis}
\end{figure}

In Figure~\ref{fig:3dnoise} and Figure~\ref{fig:3dnoise_vis}, we illustrate the framework of generating our ``3D-aware'' noise.  
Specifically, we build a 3D convolution-based novel view synthesis pipeline (denoted as $\mathcal{S}$), trained as a multi-view image reconstruction task. Similar to~\cite{wang2021facevid2vid} and~\cite{Siarohin_2019_NeurIPS},  we first employ a 3D appearance feature extraction
network to map the reference image to a 3D appearance feature volume. To synthesize an image at a novel view, we follow the volume rendering as in NeRF~\cite{mildenhall2020nerf} to integrate the 3D features into a 2D feature map which is further decoded to the final RGB image $\tilde{I}$ with a deep 2D convolutional network. The network modules are trained with image reconstruction losses against ground-truth multi-view images, including pixel-aligned L$_2$ and VGG perceptual losses~\cite{johnson2016perceptual}. 

During inference, given a reference image, we first employ our trained 3D novel view synthesis network $\mathcal{S}$ to generate a proxy rendering $\tilde{I}$ at the target view. While being blurry, $\tilde{I}$ contains rich 3D structural semantics and acts as a good guidance to the diffusion process in our DiffPortrait3D. 
We incorporate this 3D awareness by generating the starting noise using the forward noising process of 1000 steps applied to the latent map of $\tilde{I},$ i.e., $\mathcal{E}(\tilde{I}).$ Better reconstruction and consistency are observed with our proposed 3D-aware noise, as evidenced in Table~\ref{tab: 3d-aware noise} numerically and in Figure 5 of the main paper visually.




\begin{figure}[h]
	\centering
	\includegraphics[width=0.95\linewidth]{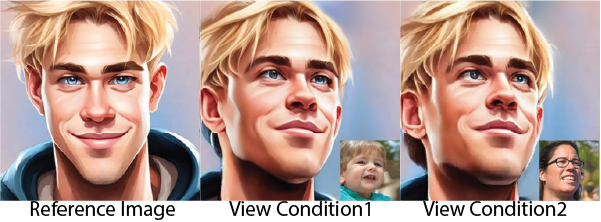}
	\centering
	\caption[Caption for LOF]{Ablation on view conditional images.}  
        \vspace{-4mm}
	\label{fig:viewcond}
\end{figure}
\begin{figure}[h]
	\centering
	\includegraphics[width=1.0\linewidth]{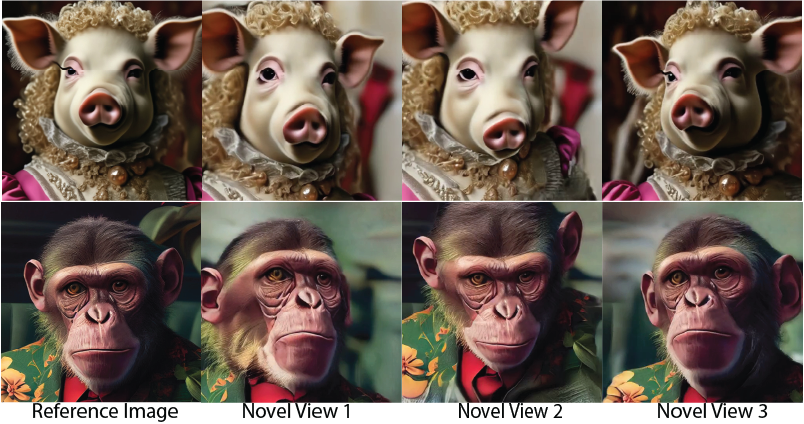}
	\centering
	\caption[Caption for LOF]{ Novel view synthesis of anthropomorphic animals.}  
        \vspace{-4mm}
	\label{fig:animal}
\end{figure}

\subsection{Metrics}
\label{sec:analysis}
\subsubsection{Identity Similarity}
Our identity similarity score (ID) is calculated based on the cosine similarity of the face embeddings with a pre-trained face recognition module\cite{deng2019arcface} as , 
\begin{equation}
    ID = {f_{g'} \cdot f_g}
\end{equation}
where $f_{g'}$, and $f_g$ are the feature embedding of the generated image and ground-truth image respectively.

\begin{figure*}
	\centering
	\includegraphics[width=\linewidth]{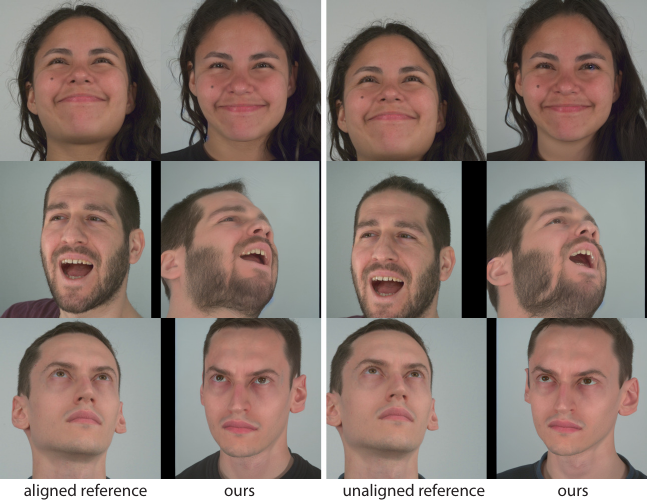}
	\centering
	\caption[Caption for LOF]{DiffPortrait3D effectively derives appearance features from the reference image, without strict restriction to its image alignment. Similar novel view synthesis results are achieved using EG3D-aligned and non-aligned reference images.}  
        \vspace{-5mm}
	\label{fig:align}
\end{figure*}
\subsubsection{Pose Accuracy}
We evaluate the pose accuracy (POSE) with the assistance of an off-the-shelf face reconstruction model \cite{deng2019accurate}. We detect pitch, yaw, and roll from the generated novel view images, then compute the L$_2$ loss against the camera poses estimated from ground truth images.

\subsection{Baselines}

\subsection{EG3D-Pivot Tuning Inversion}
For our baseline EG3D-PTI, we follow the standard procedure as described in  \cite{roich2022pivotal}, where for each reference image, we first optimize the latent noise for 500 iterations and further fine-tune the generator weights for another additional 250 iterations. Once completed, we used the optimized latent noise finetuned 3D-aware generator to synthesize the image at novel views. 

\subsection{Zero-1-to-3}
Zero-1-to-3~\cite{liu2023zero1to3} is one of the state-of-the-art novel diffusion-based view synthesis works designed for general 3D objects. Nevertheless, we compare to it for a thorough evaluation of existing works on novel portrait synthesis. In Table~\ref{tab: comparsion zero1to3}, to maximize its performance on our task, our numerical results are all based on portraits with removed backgrounds. Additionally, due to the differences in 3D camera coordinates, we only report the FID and identity similarity (ID) of the novel view synthesis results for a fair comparison.  
\begin{table}
\centering
\begin{tabular}{ccc}

    \toprule
     &3D-Aware noise 
     &Random noise
     \\
    \midrule
    LPIPS $\downarrow$ & \textbf{0.27} & 0.32   \\ 
    \midrule
    SSIM $\uparrow$& \textbf{0.68} & 0.65  \\ 
    \midrule
    DIST $\downarrow$ & \textbf{0.18} & 0.21  \\     
    \midrule
    ID $\uparrow$ & \textbf{0.70} & \textbf{0.70} \\  
    \midrule
    FID $\downarrow$ & \textbf{25.37} & 26.81 \\ 
   \bottomrule
\end{tabular}
\caption{Quantitative ablation of 3D-Aware Noise and Random Noise results of novel view synthesis of NeRSemble~\cite{kirschstein2023nersemble} at the resolution of 512x512}
\label{tab: 3d-aware noise}
\end{table}
\section{More Experiment results}
\label{sec:moreres}

\subsection{Alignments}
Our model was trained with EG3D-aligned reference and target images. However, our method does not restrict the reference images to be cropped and aligned, nor with the camera condition images.
In Figure~\ref{fig:align}, we showcase that with differently aligned reference images, our method synthesizes close novel view results. 

\subsection{PanoHead-PTI Comparison}
\begin{figure}[h]
	\centering
	\includegraphics[width=1.0\linewidth]{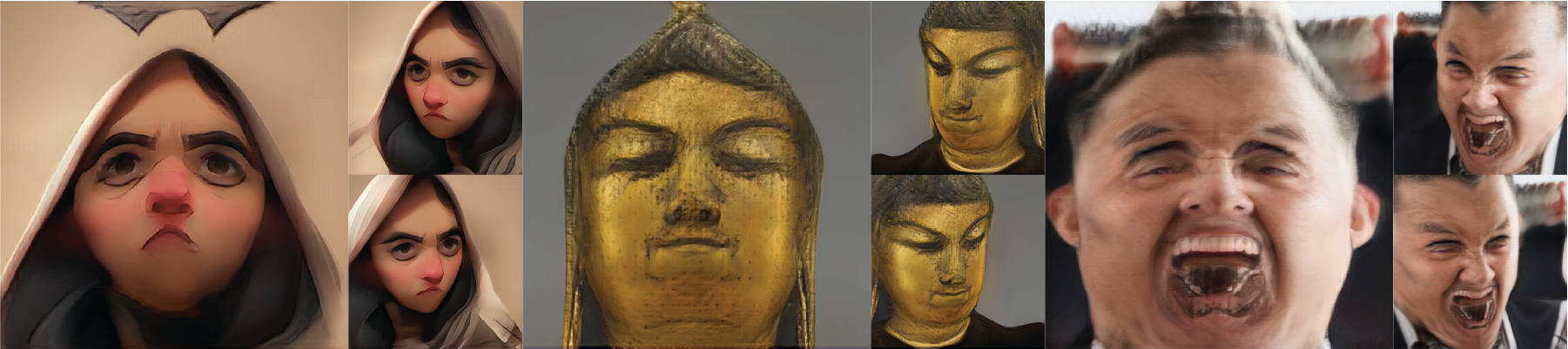}
	\centering
	\caption[Caption for LOF]{Novel view synthesis with PanoHead-PTI.}  
        \vspace{-4mm}
	\label{fig:panohead}
\end{figure}

PanoHead extends the EG3D framework by enabling novel view synthesis in 360$^\circ$. However, owing to the inherent limitations of GAN-based architecture, PanoHead, like EG3D, necessitates time-consuming instance-specific optimization (pivot-tuning) while still suffers from limited perceptual quality and identity loss, especially for portraits with out-of-domain styles or extreme expressions (as shown in Fig.~\ref{fig:panohead} compared to our results in Fig. 4 of the main paper). Our method is superior by a large margin on quantitative metrics as well (POSE$\downarrow$-/0.0023/-, LPIPS $\downarrow$ 0.22/0.28/0.11, SSIM $\uparrow$ 0.60/0.53/0.76, DIST $\downarrow$ 0.18/0.26/0.12, ID$\uparrow$ 0.47/0.38/0.12, FID $\downarrow$ 56.53/60.4/90.47; ours are detailed in Tab.1 of the main paper). 

\begin{table}
\centering
\begin{tabular}{ccc}

    \toprule
     &Ours
     &Zero1to3
     \\
    \midrule
    LPIPS $\downarrow$ & \textbf{0.04/0.19/0.04} & 0.28/-/0.34  \\ 
    \midrule
    SSIM $\uparrow$& \textbf{0.90/0.74/0.88} & 0.70/-/0.52   \\ 
    \midrule
    DIST $\downarrow$ & \textbf{0.05/0.15/0.04} & 0.16/-/0.19  \\     
    \midrule
    ID $\uparrow$ & \textbf{0.94/0.70/0.92} & 0.82/0.09/0.70 \\  
    \midrule
    FID $\downarrow$ & \textbf{6.5/32.6/11.6} & 61.4/113.8/57.5 \\ 
   \bottomrule
\end{tabular}
\caption{ Quantitative comparison of our method and Zero-1-to-3, showing numerical results of reconstruction/novel view synthesis of NeRSemble~\cite{kirschstein2023nersemble}, and reconstruction of in-the-wild test images( from left to right). For a fair comparison to Zero-1-to-3, the evaluation is performed with the removed backgrounds at the resolution of $512\times 512$.}
\label{tab: comparsion zero1to3}
\end{table}

\subsection{View-consistent novel view synthesis}
 We show more challenging results in Figure~\ref{fig:more_vis1}
,~\ref{fig:more_vis2} and ~\ref{fig:more_vis3}. Our model is able to generalize well to arbitrary face portraits with unposed camera views, extreme facial expressions, and diverse artistic depictions. Please also refer to our supplementary video for more high-resolution results.

\subsection{Ablation on view condition images}

Our method effectively disentangles the control of camera views from appearance. As shown in Fig.~\ref{fig:viewcond}, visual differences are hardly noticeable between the synthesized novel views when using two view conditions generated at the same camera pose but with distinct appearance seeds. For quantitative assessment, we perform novel view synthesis across all our test images using two sets of   view conditional images generated under the same camera pose but featuring different appearances. We calculate the differences in image pixels (LPIPS $\downarrow$ 0.09) and camera poses (POSE$\downarrow$ 0.0041). Note that the LPIPS difference is partially attributed to slight structural shifting.

\subsection{Anthropomorphic animals}  While being trained sorely on real human images, our method is empowered with strong domain generalization capability (e.g., Fig.~\ref{fig:animal}) by leveraging the generative prior of a pre-trained stable diffusion model. However, we acknowledge that visual artifacts are possible due to the appearance bias originating from the training data distribution.

\section{Limitation and Future Work}
\label{sec:limit}
\begin{figure*}
	\centering
	\includegraphics[width=\linewidth]{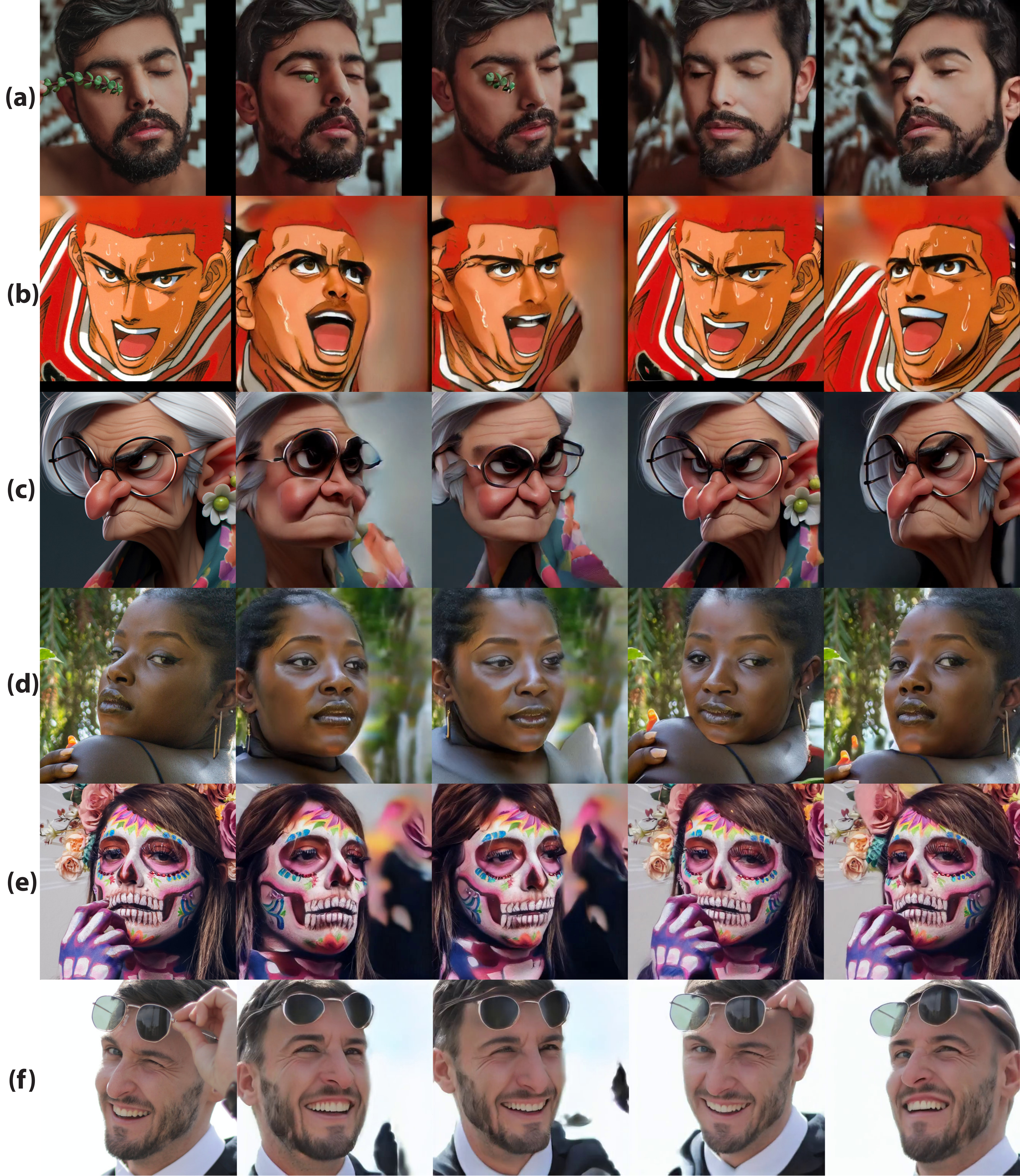}
	\centering
	\caption[Caption for LOF]{Failure Case}  
	\label{fig:fail_case}
\end{figure*}

\begin{figure*}
	\centering
	\includegraphics[width=\linewidth]{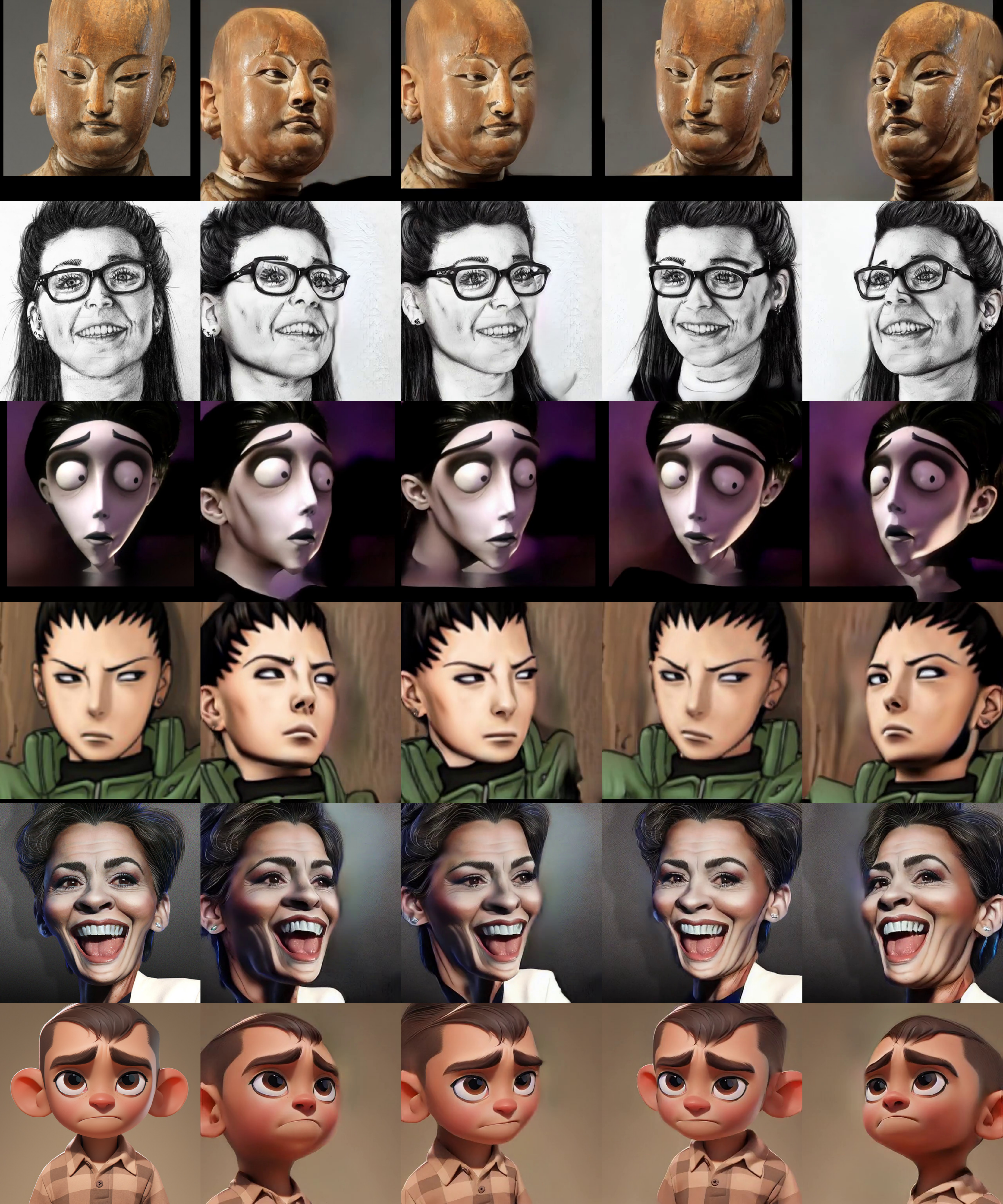}
	\centering
	\caption[Caption for LOF]{More Novel-view consistent results}  
	\label{fig:more_vis1}
\end{figure*}

\begin{figure*}
	\centering
	\includegraphics[width=\linewidth]{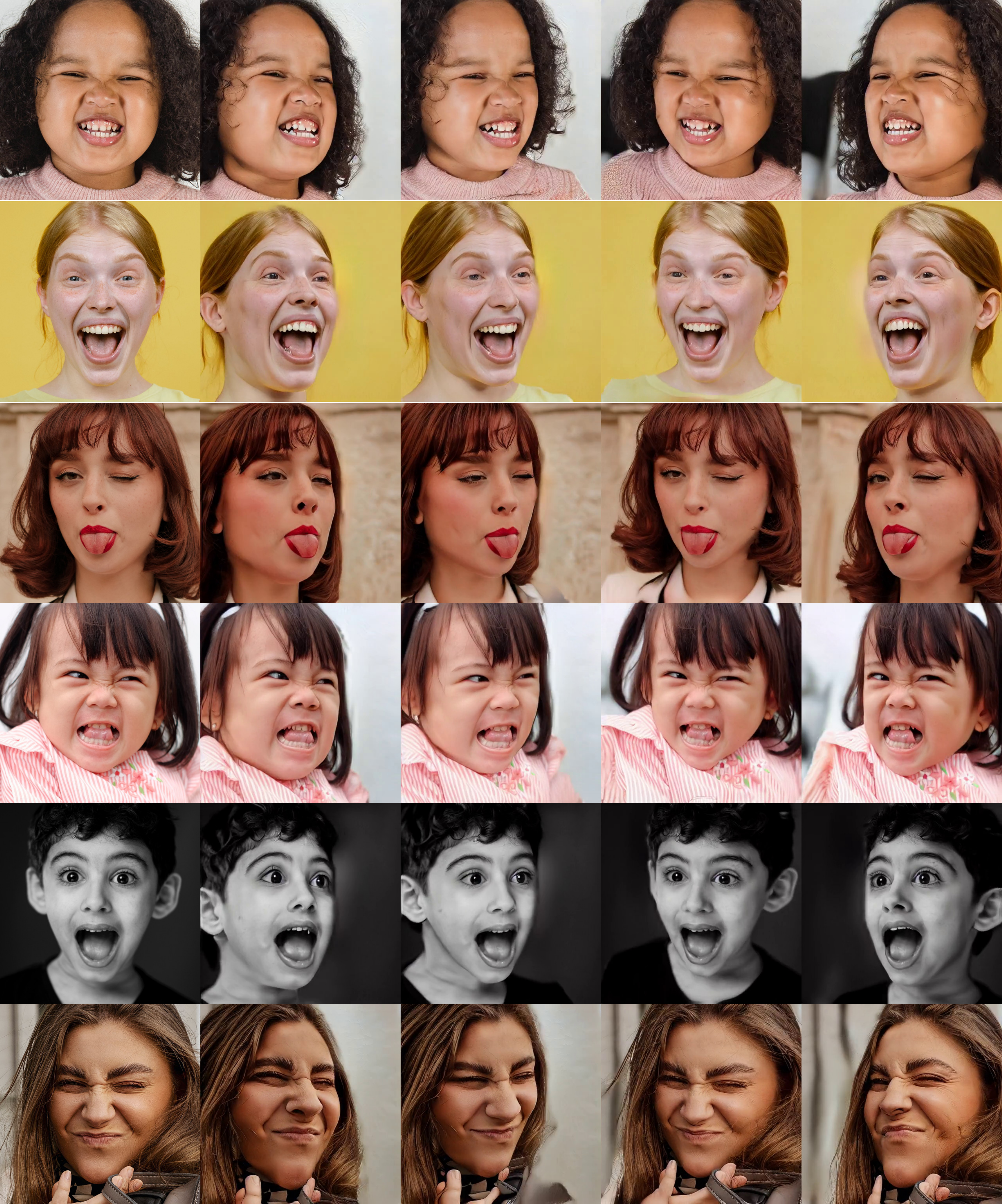}
	\centering
	\caption[Caption for LOF]{More Novel-view consistent results}  
	\label{fig:more_vis2}
\end{figure*}

\begin{figure*}
	\centering
	\includegraphics[width=\linewidth]{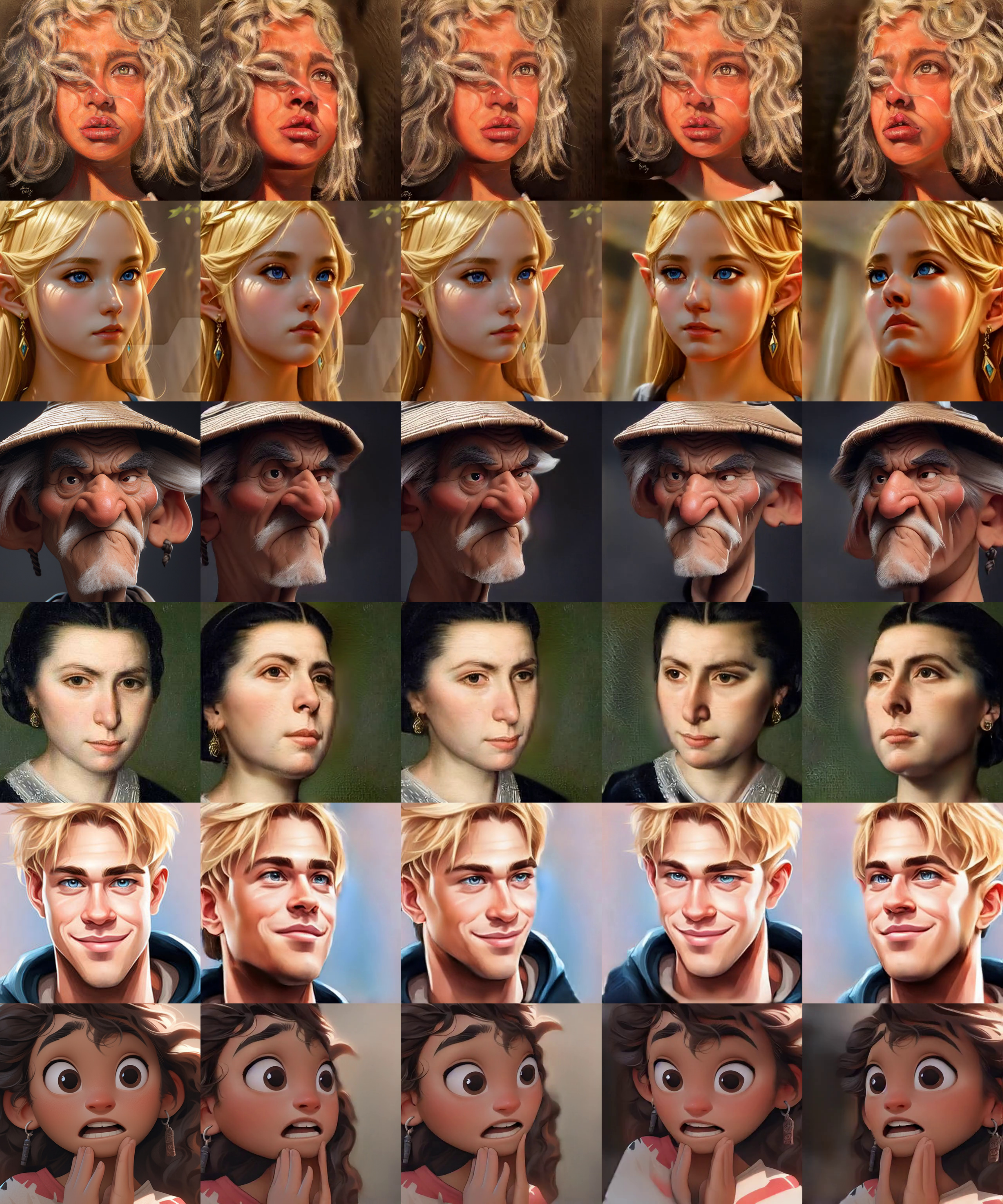}
	\centering
	\caption[Caption for LOF]{More Novel-view consistent results}  
	\label{fig:more_vis3}
\end{figure*}

While the image coherence is largely strengthened with our cross-view module and 3D-aware generation, we still observe occasionally flickering artifacts in unobserved regions. We leave the exploration of longer-range consistent view manipulation as future work. 
In this work, the appearance is formulated to be sourced from the reference images only. This could result in some loss of identity given the limited appearance context. In the future, we would like to extend our framework such that the identity can be multi-sourced from e.g., text and personalized Loras~\cite{hu2021lora}. 
As discussed above, we also include visualizations of failure cases in Figure~\ref{fig:fail_case}. Rows (a) through (f) display artifacts in areas not observed, accompanied by changes in identity, particularly noticeable in (b), (c), and (d). In cases (a), (e), and (f), it is evident that the model struggles to accurately replicate secondary elements from the reference image, such as sunglasses in (f), hands and flowers in (e), and leaves in (a), leading to inconsistent outcomes. One potential issue we've identified stems from the use of a 2D diffusion backbone that integrates 3D-Aware information. This approach, while innovative, may lead to minor inconsistencies and deviations, especially in challenging depictions. Addressing these limitations is an important area that should be addressed in future work.

\section{Ethic Consideration}
\label{sec:ethic consider}
We acknowledge the profound capabilities of the diffusion model as a powerful generative model. The framework proposed in our paper could, theoretically, be utilized to compromise multi-perspective facial recognition systems. We assert that the model and the accompanying research code are intended exclusively for advancing scientific research and must not be used for illicit purposes.

{
    \small
    \bibliographystyle{ieeenat_fullname}
    \bibliography{main}
}


\end{document}